\documentclass[10pt,twocolumn,letterpaper]{article}

\usepackage[pagenumbers]{cvpr} 

\usepackage{algorithm}
\usepackage[noend]{algpseudocode}
\usepackage{amssymb}
\usepackage{amsmath}

\DeclareMathOperator*{\argmin}{arg\,min}

\definecolor{cvprblue}{rgb}{0.21,0.49,0.74}
\usepackage[pagebackref,breaklinks,colorlinks,allcolors=cvprblue]{hyperref}

\def\paperID{10462} 
\def\confName{CVPR}
\def\confYear{2025}

\title{Light Transport-aware Diffusion Posterior Sampling \\for Single-View Reconstruction of 3D Volumes}

\author{Ludwic Leonard\\
{\tt\small ludwig.mendez@tum.de}
\and
Nils Th\"urey\\
{\tt\small nils.thuerey@tum.de}
\and
R\"udiger Westermann\\
{\tt\small westermann@tum.de}
\and
Technical University of Munich
}

\begin{document}
\twocolumn[{%
\renewcommand\twocolumn[1][]{#1}%
\maketitle
\begin{center}
    \centering
    \captionsetup{type=figure}
    \includegraphics[width=\textwidth]{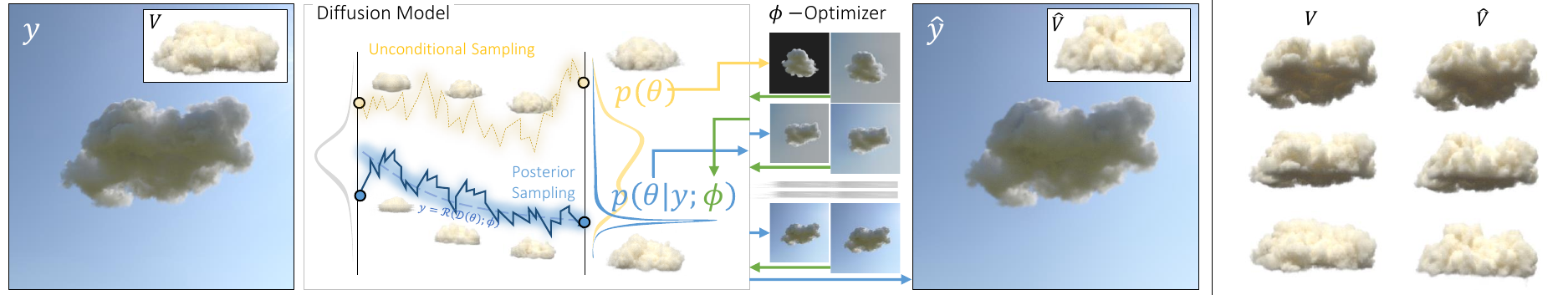}
    \captionof{figure}{
    Given a single view ($y$) of a volume ($V$), we reconstruct a volume ($\hat{V}$) from its latent representation ($\theta$) that matches $y$ under the same lighting conditions, resulting in a synthesized view ($\hat{y}$). A differentiable volume renderer ($\mathcal{R}$) is used to optimize physical scene parameters ($\phi$) while simultaneously performing posterior sampling $p(\theta|y;\phi)$, conditioned on the observation, in the latent space of a trained diffusion model $p(\theta)$. Ambiguities due to the absence of information about unseen parts of the volume are reduced by gradually steering the reverse diffusion process toward the most plausible reconstruction under the given view (right section). 
    }
    \label{fig:teaser}
\end{center}%
}]
\begin{abstract}
We introduce a single-view reconstruction technique of 
volumetric fields in which multiple light scattering effects are omnipresent, such as in clouds. 
We model the unknown distribution of volumetric fields using an unconditional diffusion model trained on a novel benchmark dataset comprising 1,000 synthetically simulated volumetric density fields.
The neural diffusion model is trained on the latent codes of a novel, diffusion-friendly, monoplanar representation.
The generative model is used to incorporate a tailored parametric diffusion posterior sampling technique into different reconstruction tasks.
A physically-based differentiable volume renderer is employed to provide gradients with respect to light transport in the latent space. This stands in contrast to classic NeRF approaches and makes the reconstructions
better aligned with observed data. 
Through various experiments, we demonstrate single-view reconstruction of volumetric clouds at a previously unattainable quality.

\end{abstract}    
\section{Introduction}
\label{sec:introduction}

The reconstruction of a 3D model from a single image \cite{dou2023tore,jun2023shap,liu2023zero,bhat2021adabins,melas2023realfusion} 
is a fundamental task in 3D computer graphics and vision. Once the model is reconstructed, operations such as novel view synthesis, relighting or inpainting can be applied. However, this problem is ill-posed and, in general, requires additional views to constrain the object parameters and infer plausible reconstructions of unseen parts.

Differentiable rendering (DR) leverages a rendering process with gradients, making it suitable for recovering shape and optical material parameters from images \cite{kato2020differentiable,wang2021from,mescheder2022gendr,durvasula2023distwar}.
DR enables backpropagation of gradients of a loss in image space to the scene parameters, including position, texture, lighting, shape, and other attributes. The challenge increases significantly when these parameters describe complex distributions of volumetric materials, such as clouds, smoke, or fire. In such scenarios, the problem becomes so ill-posed that it requires dozens, if not hundreds, of different views to adequately constrain the object parameters ~\cite{nimier2019mitsuba,nimier2020radiative,leonard2024image}. It is now widely accepted that reconstructing the internal density distribution of highly dense volumes is nearly impossible due to the high uncertainty in the light scattering process and the presence of vanishing gradients. This limitation can only be alleviated by incorporating prior information during reconstruction.

When sufficient 3D datasets representing different instances of an object type are available, network inference can be used to tackle the task of inferring the 3D object. Many recent approaches build upon generative diffusion models that are trained on 3D datasets \cite{luo2021diffusion,nichol2022point,gao2022get3d,liu2023meshdiffusion, zhang2023holofusion,gupta2023hyperdiffusion,zhou2023sdfusion,anciukevicius2023renderdiffusion,muller2023diffrf}. 
Diffusion models have gained popularity for their ability to produce high-quality, realistic 3D samples of specific object categories. 

Using diffusion models for single-view volume reconstruction, however, is challenging. Firstly, a publicly available 3D dataset on which a diffusion model can be trained is not existing. Secondly, an image that is taken in the wild contains intricate illumination effects due to background light and multiple light scattering in the volume interior. While the scattering properties of the material can be assumed, the background radiance is typically unknown but needs to be inferred to separate the object. 
In general, if the optical parameters are not resolved, it is impossible to understand how the appearance is explained. 

Our proposed approach addresses these challenges by employing a diffusion prior to guide a Physically-based Differentiable Volume Renderer (PDVR) toward reconstructing a plausible volumetric field. 
In contrast to previous approaches, 
our approach includes controlled variations in the diffusion step by considering the gradient of the image loss with respect to the learned latent space representation. This approach steers the reconstruction toward a realistic 3D density distribution, ensuring that the generated structure aligns well with the observed data and maintains realistic spatial consistency.

The diffusion model is trained on a dataset comprising 1,000 synthetically simulated volumetric density fields (specifically, cumulus clouds in our case study), using a novel, diffusion-friendly representation for decoding. The reconstruction is simultaneously constrained by the diffusion prior and the image containing light transport effects. The renderer is coupled with the diffusion model to reconstruct radiance parameters using the prior for the density distribution but not for its appearance. Thus, the 3D density field can also be trained solely on the prior, not requiring images of all possible backgrounds and light scattering effects.

Our key contributions are as follows:

\begin{itemize}
    \item A large database of 3D cumulus cloud-like density fields, generated using numerical fluid simulation.
    \item A 3D cloud decoder utilizing a novel, diffusion-friendly monoplanar representation, trained jointly on a subset of the database.
    \item A novel Parametric Diffusion Posterior Sampling (PDPS) technique utilizing a shape-centric prior with a physically-based differentiable volume renderer.
\end{itemize}

To the best of our knowledge, this is the first approach that integrates an unconditional diffusion model, trained on volumetric density distributions, with a differentiable volume renderer. We demonstrate the potential of our approach across various tasks, including single- and multi-view reconstruction, and volume super-resolution.

\section{Related Work}

\paragraph{3D model reconstruction for view synthesis} Novel view synthesis aims at computing a 3D scene representation from 2D input images of this scene, and uses this representation to generate novel views from arbitrary viewpoints. NeRF-style approaches \cite{mildenhall2021nerf} learn a 3D Neural Radiance Field (NeRF) which can be rendered with direct volume rendering. A number of techniques have recently been proposed to make NeRF fast and scalable in the size of the features it can reconstruct \cite{yu_plenoctrees_2021,fridovich-keil_plenoxels_2022,mueller2022instant,turki2022meganerf,tancik2022blocknerf}. 

NeRFs have been generated initially with MLP-based Scene Representation Networks (SRNs) \cite{sitzmann_scene_2019}, which have later been used to compactly encode volumetric scalar fields using the emission-absorption optical model ~\cite{lu2021compressive,weiss2022fast}. Alternative to the use of SRNs, adversarial approaches have recently emerged. They use 2D images to stochastically condition the 3D reconstruction using an adversarial loss \cite{schwarz2020graf,chan2021pigan,gu2021stylenerf,niemeyer2021giraffe,zhou2021cips3d}. In this context, sparse tri-plane volumetric models have been proposed to reduce the memory consumption at improved training efficiency of NeRFs \cite{fu2023threedgen,chen2022tensorf}. While NeRF-based approaches usually assume that images of the scene from many different viewpoints exist, recent advancements have shown their potential to also perform single-view reconstruction \cite{chan2021pigan, wang2021ibrnet,schwarz2020graf,mueller2022autorf,yu2021pixelnerf,lin2023vision}.  

\paragraph{Generative diffusion modeling}
Generative diffusion modeling \cite{sohl2015deep} has paved the way for what nowadays is termed “diffusion models” \cite{score_matching2020,song2020improved,ddim2021,song2021score,ho2020denoising}, i.e., the creation of synthetic data, such as images, audio, and text, by iteratively refining random noise into structured outputs. Karras et al. ~\cite{po2023state} and Po et al. ~\cite{po2023state} provide thorough overviews of the current research in this field. For 3D reconstruction tasks, the diffusion model, i.e., the latent (compressed) space, is used as a generative prior for the underlying structure and features of the data. Previous works focus on the reconstruction of purely geometric representations \cite{zeng2022lion,nichol2022point,luo2021diffusion,zhou2021shape}, neural fields \cite{zhou2023sdfusion,gupta2023hyperdiffusion,melas2023learning,fu2023threedgen,jun2023shape,zhang2023holofusion,kim2023neuralfield,chen2023singlestage,muller2023diffrf,ntavelis2023autodecoding,zhang2023shape} or use 2D image diffusion models to generate 3D models, either directly or via factorized radiance representations \cite{poole2022dreamfusion,bautista2022gaudi,wang2023rodin,chen2022tensorf,li2022threedesigner}. Generative diffusion models have been used for single-view 3D reconstruction, either for novel view synthesis without an underlying geometric model \cite{watson2022novel,liu2023zero,jain2023genvs}, or by computing this model iteratively aside of the denoising process \cite{schwarz2020graf,mueller2022autorf,yu2021pixelnerf,szymanowicz2023viewset,tewari2023diffusion,zhou2021dmv3d,long2023wonder3d}. 

Instead of performing denoising directly in the pixel or voxel space, operating in the space of a compressed latent representation \cite{rombach2022high,shue20233d,podell2023sdxl,tewari2022advances} offers considerable advantages. Once a sample from the latent representation is obtained, a decoder $\mathcal{D}(\theta)$ is used to reconstruct the final signal, such as volumes \cite{zhu2023make,zhou20213d}, signed distance fields \cite{chou2023diffusion,cheng2023sdfusion}, and radiance fields \cite{karnewar2023holodiffusion,anciukevicius2023renderdiffusion,chen2023singlestage}. This approach not only enhances efficiency but also utilizes the structured features learned within the latent space, promoting greater consistency and coherence in the final decoded output.

\paragraph{Image-based volume reconstruction} Zhang et al.~ \cite{zhang2019differential} present a general framework, including volumetric media, for calculating radiance derivatives with respect to changes of scene parameters. This framework has later been extended to make it applicable to path tracing including random sampling \cite{zhang2021path}. Properties of the differentiation of integrators are analyzed by Zeltner and Monte \cite{zeltner2021monte}. Forward mode automatic differentiation \cite{nimier2019mitsuba} for differentiable rendering is nowadays replaced by radiative backpropagation \cite{nimier2020radiative} to decrease the required memory, yet at the expense of multiple branches along light paths and quadratic time complexity thereof. Performance increases are achieved by reusing radiances along light paths \cite{vicini2021path}, and by avoiding recursive radiance estimates at scattering locations with dedicated sampling methods for estimating derivatives of volumetric scattering ~\cite{nimier2022unbiased}. 
For multi-view reconstruction of volumetric fields in the presence of global light transport, singular path sampling in combination with in-scattering relaxation and an exponential moving average shows improved reconstruction fidelity \cite{leonard2024image}. Under the assumption of an emission-absorption optical model, the ``inversion trick'' enables fast automatic differentiation for volume reconstruction and transfer function learning ~\cite{weiss2021differentiable}. Physical constraints are combined with self-supervision for the reconstruction of single-scattering flow fields from single-view videos \cite{franz_global_2021}.

\section{Problem Formulation}
\label{sec:problem}

\subsection{Diffusion Models}
A diffusion model operates by applying a forward \textit{Markov chain} process to an initial data sample \( x_0 \), gradually transforming it into pure Gaussian noise at a final state \( x_T \), where \( T \) is typically large (e.g., \( T \sim 1{,}000 \)). This transformation is governed by a fixed, time-dependent Gaussian transition distribution \( q(x_t \mid x_{t-1}) \). The model then trains a reverse Markov chain, parameterized by a set of distributions \( p_\varPhi(x_{t-1} \mid x_t) \), which also take the form of Gaussians. The training objective is usually to predict the noise \( \epsilon_t \) that was incrementally added in the forward process, enabling the model to reconstruct the original data \( x_0 \) by denoising sequentially from \( x_T \) back to \( x_0 \).

\subsection{Diffusion Posterior Sampling}
Given a forward model $y := \mathcal{A}(x_0) + \eta$, with $\eta$ assumed to be white Gaussian noise, a probabilistic model $p(y | x_0) = \mathcal{N}(y; \mathcal{A}(x_0), \Sigma)$ represents the conditional probability of obtaining the observation given some parameters $x_0$. With a prior $p(x_0)$, represented as an unconditional diffusion probabilistic model, the posterior distribution $p(x_0 | y)$ can be approximated as in DPS \cite{chung2022diffusion} using Bayes inference.

The approach aims to bypass the indirect dependency $p(y | x_t)$ that exists for all $x_t$ except $x_0$ by introducing an estimate $\hat{x_0}(x_t)$ for $x_0$ at each level.

Adding the gradient
\begin{equation}
\zeta \nabla_{x_t} \|y - \mathcal{A}(\hat{x_0}(x_t))\|^2_2    
\end{equation}
at each step guides the reverse process of an unconditional diffusion model toward the posterior sample. Here, \(\zeta\) is a hyperparameter that balances prior enforcement with observation fidelity by accounting for normalization and the noise level of the measurement (see \cite{chung2022diffusion}).

\subsection{Differentiable Rendering with a Diffusion Prior}
When measurements $y$ involve complex physical phenomena, such as light transport through a medium with multiple scattering, the process $\mathcal{A}$ must account for these complexities. A differentiable rendering process $\mathcal{R}(\phi)$ enables us to simulate these effects by modeling how light interacts with the medium (e.g., clouds, smoke) and reaches the observer or sensor. Additionally, it provides a method to compute how the gradients of a loss function with respect to the rendered image, $\nabla_{\mathcal{R}} \mathcal{L}$, propagate through all the parameters $\phi$ that govern the light scattering and interaction dynamics.

However, differentiable volume rendering faces challenges in accurately reconstructing scene parameters when limited to only a few input images, as the optimization process may not have enough information to fully constrain the volume’s density distribution and material properties. Therefore, our goal is to learn a volumetric prior that synthesizes plausible cloud-like density fields via a diffusion model. Since such models struggle to generalize or precisely reconstruct details of objects or configurations that were not included in their training data, our key problem is how to embed the volume prior into a differentiable volume renderer ensuring that the generated structure aligns well with observed data and maintains realistic spatial consistency.

\section{Method}

To address the problem formulated in Section~\ref{sec:problem}, we propose a diffusion posterior sampling scheme in combination with a differentiable volume renderer to simultaneously consider physical light transport effects in a single view and a cloud-aware prior. Figure \ref{fig:teaser} provides an overview of our method. Starting from a synthetically generated cumulus cloud database (see Section~\ref{sec:cloudy_ds}), our posterior sampling scheme employs a latent diffusion model to generate a 3D density field with characteristic cloud distribution (see Section~\ref{sec:volume_latent_space}).

\begin{figure}[t]
    \centering
    \includegraphics[width=0.9\linewidth]{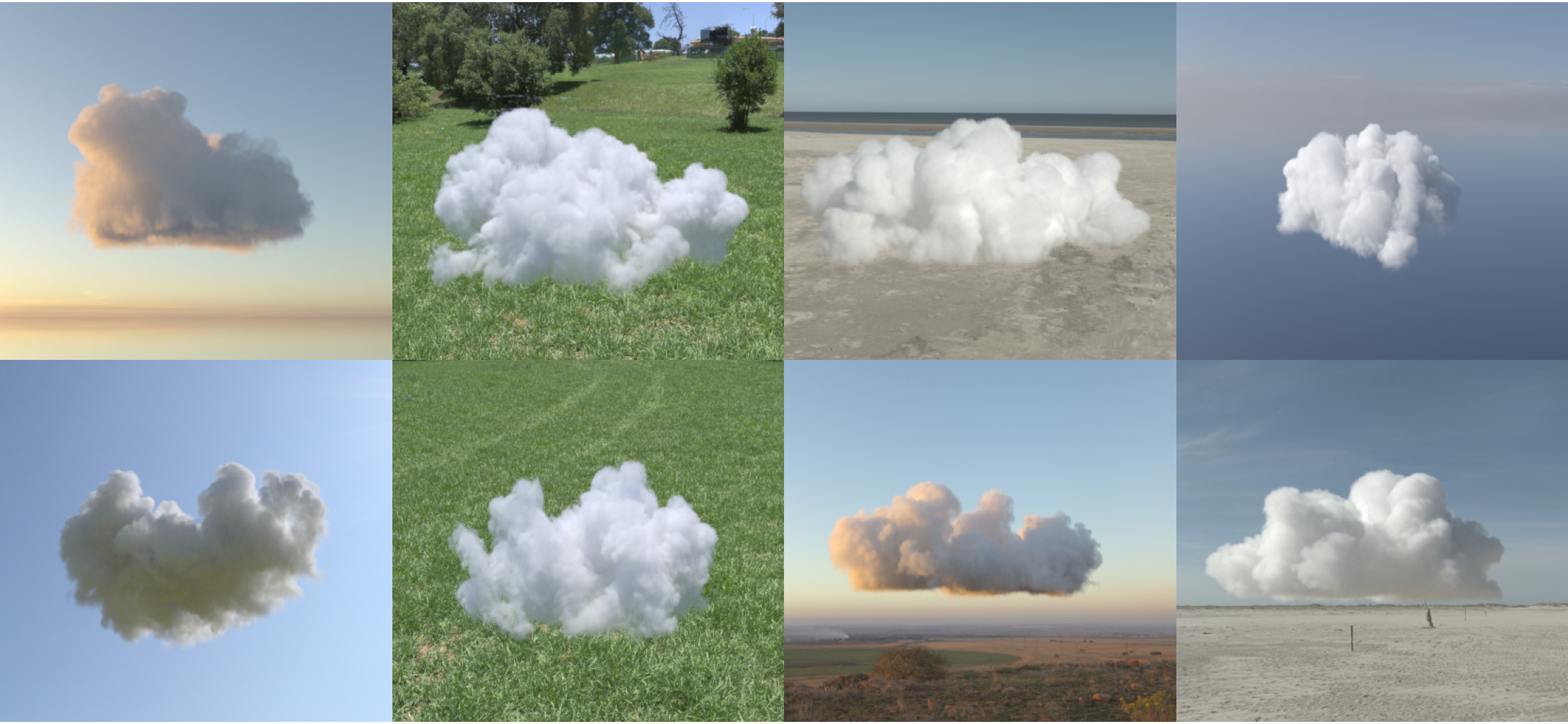}\\
    \vspace{0.4em}
    \includegraphics[width=0.9\linewidth]{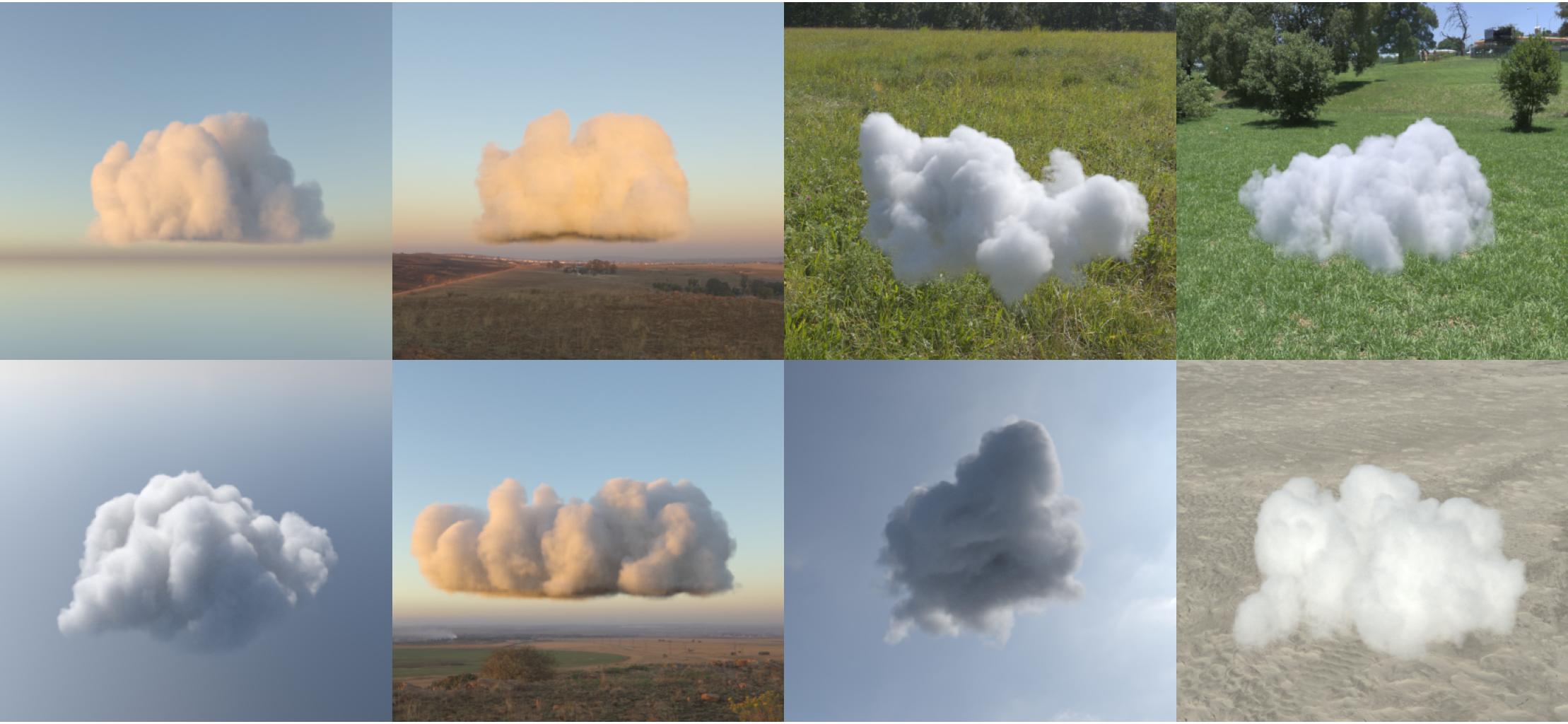}
    \caption{Top images: Cloudy Dataset -- Photorealistic renderings of randomly selected clouds from our dataset, illustrating natural variations and details. Bottom images: Diffusion-based cloud synthesis -- Clouds generated with our diffusion model, demonstrating a convincing appearance under realistic lighting conditions and physical parameters.
    }
    \label{fig:dataset_generation}
\end{figure}

We introduce our novel monoplanar latent representation to effectively compress the cloud database (see Section~\ref{sec:method_monoplanar}), and we demonstrate how to prevent overfitting by refining this latent representation through analog transformations in both spatial and latent space (see Section~\ref{sec:volume_latent_space}). With a standard volume diffuser reconstructing a cloud by sampling from the latent representation, we constrain the reverse Gaussian process to a parameterized posterior sample (see Section~\ref{sec:method_pdps}).

Finally, differentiable volumetric path-tracing \cite{leonard2024image} with Monte Carlo importance sampling is used to account for the recursive dependency of the incoming radiance at scattering positions, iterating over all possible path lengths. The diffuser serves as a prior for a subset of recovered scene parameters (see Section \ref{sec:optimization}).

\subsection{Cloudy - a 3D Clouds Dataset}
\label{sec:cloudy_ds}
First, we create a dataset consisting of 1,000 synthetic clouds using the JangaFX fluid simulator \cite{jangafx2024}. The simulator is  configured to emulate the evolution and dynamics of gaseous substances, capturing realistic buoyancy, turbulence, and diffusion essential for producing the lifelike flow and rising motion characteristic of vapor and cloud formation.

To add natural randomness and represent diverse distributions of warm columns to the clouds, we apply  Perlin noise functions and varied particle emission shapes. Figure \ref{fig:dataset_generation} (top) shows a random selection of clouds from our dataset, which are rendered under different lighting conditions. The density fields are numerically simulated on regular 3D grids at a resolution of approximately \(512 \times 256 \times 512\).

\subsection{Volume Latent Encoding}
\label{sec:method_monoplanar}

We introduce an implicit neural representation for a volume $\mathcal{V}$ defined on the cube $[-1,1]^3$, based on a single projection, which we refer to as \emph{monoplanar}. Unlike previous approaches that use positional feature embeddings like triplane or tensor decomposition, our method involves sampling a window across a single projected axis, centered at the coordinate of interest, to extract the final features.

\begin{figure}
    \centering
    \includegraphics[width=\linewidth]{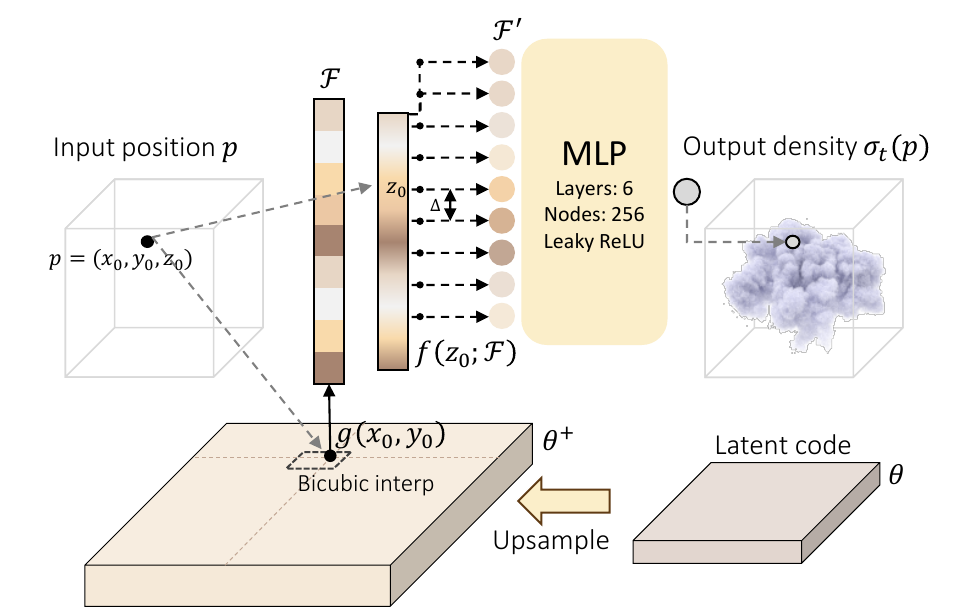}
    \caption{Implicit monoplanar representation.}
    \label{fig:irn}
\end{figure}

Let $g: \mathbb{R}^2 \rightarrow \mathbb{R}^N$ be a continuous two-dimensional field of features based on a grid, i.e., $g(x,y)$ returns a $1$-dimensional vector with $N$ sampled values using bicubic interpolation. The vector is structured as another grid $\mathcal{F}$ with domain $[-1,1]$. The function $f(z; \mathcal{F})$ samples $\mathcal{F}$ at positions $z - 1 + k*\Delta, k \in \{0 \dots N-1\}, \Delta = 2 / (N-1)$ using linear interpolation. Sampled positions are constrained to $[-1,1]$. The feature vector $\mathcal{F'}$ storing the $N$ interpolated samples is fed into an MLP to produce the final density value, see Figure \ref{fig:irn}.

In practice, we parameterize $g$ with a coarse grid $\theta$ of size $128\times128$ and $32$ features. A convolutional upsampler is applied to increase the resolution to $256\times256 \times64$. Once upsampled, the feature vector at a specific position $(x_0,y_0,z_0)$ is obtained using $g$ and $f$ described earlier. 

The monoplanar representation model is trained jointly on a subset of the clouds from the Cloudy dataset, sharing the parameters for the upsampler and the MLP decoder. This approach is common in triplanar-based 3D generative models \cite{chen2023singlestage, fu2023threedgen, shue20233d}.
We found that $64$ cloud samples are sufficient to obtain an accurate latent encoding.
Thus, only the parameters of the latent grid $\theta$ are representative of the volume. The representation is constrained to be equivariant to flips and transpositions of the latent grid. The final latent code is about $2\textsc{MB}$. Since the memory consumption of a single cloud is roughly $100\textsc{MB}$, this results in a 50x compression.

While, in theory, the implicit representation $\mathcal{V}(\cdot; \theta)$ encoded in an MLP could be queried directly within a differentiable renderer, we opt to use a proxy grid $\mathcal{D}(\theta)$ that explicitly exposes all volume values. A grid only requires trilinear interpolation on the GPU, making it easier to integrate and evaluate in a differentiable renderer. Gradients of the grid can be backpropagated through the model after they are computed. 

\subsection{Volume Latent Space}
\label{sec:volume_latent_space}

To effectively train a diffusion model, it is essential to sufficiently cover the entire data manifold. Training with only a few instances would lead to a tendency for overfitting, limiting the model's ability to generalize features for unseen clouds.

To generate the space of latent representations used to guide the reconstruction process, we consider all 1,000 clouds from the Cloudy dataset and generate the respective latent codes by optimizing the decoder \(\mathcal{D}(\theta)\) using gradient descent.

\begin{figure}[t] 
    \centering
    \includegraphics[width=\linewidth]{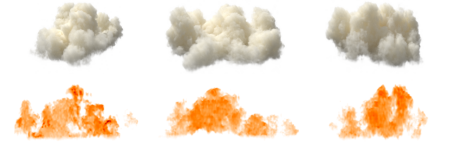}
    \caption{
Diffusion Sampling. First column: A cloud from the Cloudy dataset. Subsequent columns show clouds generated by our diffusion model. First row shows the clouds under neutral
lighting conditions, demonstrating realistic cloud-like formations. Bottom row shows cross-sectional slices through the volumes, demonstrating realistic interiors of diffused clouds.}
    \label{fig:generating}
\end{figure}

Since cloud formations are equivariant to arbitrary rotations and minor scaling along the $xy$-plane, we apply 14 such operations to the clouds and augment the dataset by these instances. The analog transformations are applied to the latent codes as an initial solution, which is then subsequently refined via optimization. While the transformed latent already represents a plausible volume, the refinement prevents the diffuser from learning patterns that emerge purely from resampling, i.e., due to boundaries and clamping (see the supplementary material for an example).
Including the $8$ equivariant transformations (flips and transposes), we obtain a total of \(1,000 \times 14 \times 8\) volume instances for training. Figure \ref{fig:dataset_generation} (bottom) demonstrates the effectiveness of our diffuser in generating new, unconditional volumetric instances. The ability to produce clouds with realistic shape and interior is demonstrated in Figure \ref{fig:generating}. 

\subsection{Parameterized Posterior Sampling}
\label{sec:method_pdps}

Let us now assume that a proper posterior sampling method $p(\theta | y; \phi)$ is available, meaning that given an observation $y$ and a forward model $y=\mathcal{A}(\theta; \phi) + \eta$, we can draw samples $\theta$ that satisfy the observation. In our case, $\mathcal{A}(\theta; \phi)$ encapsulates both the decoding of the volume from $\theta$ and the rendering depending on $\phi$, i.e., $\mathcal{A}(\theta; \phi) := \mathcal{R}(\mathcal{D}(\theta), \phi)
$.

The parametrization $\phi$ refers to unknown parameters, independent of $\theta$ which may govern other aspects of the rendering, such as environmental settings, density scales, phase functions, and scattering albedos.

With this setup, the reconstruction of all parameters $\phi$ and $\theta$ can be obtained by optimization with respect to the following objective:
\begin{equation}
\hat{\phi} = \argmin_\phi \mathbb{E}_{p(\theta | y; \phi)} \left[ \|y - \mathcal{A}(\theta;\phi)\|^2_2 \right],
\label{eq:fulloptim}
\end{equation}
where the expectation is taken over the posterior distribution $p(\theta | y; \phi)$. 

The optimization is performed with Stochastic Gradient Descent (SGD). The parameters $\phi$ are updated each step using the gradients of the argument in (\ref{eq:fulloptim}) estimated with a single sample $\theta$ as 
\begin{equation}
\label{eq:phi_gradient}
\nabla_\phi \|y - \mathcal{A}(\theta; \phi)\|^2_2.
\end{equation}

After determining $\hat{\phi}$, the final latent representation $\theta$ can be sampled from the posterior distribution $\theta \sim p(\theta | y; \hat{\phi})$. In addition to the loss in Eq.~\ref{eq:fulloptim}, we can incorporate a regularization term \( \mathcal{L}_{\textsc{reg}}(\phi) \) to enforce additional priors on the physical parameters.

\subsection{Optimization}
\label{sec:optimization}

A naive application of SGD to (\ref{eq:fulloptim}) is impractical due to the high computational cost associated with evaluating $p(\theta|y; \phi)$. This process requires computing $\nabla_{x_t} \|y - \mathcal{A}(\hat{x}_0(x_t); \phi)\|^2_2$ thousands of times. 

Depending on the complexity of $\mathcal{A}(\theta; \phi)$ with respect to the parameters, it may be advantageous to reuse the same sample $\theta$ for multiple steps in a pass, during the overall optimization. This strategy reduces the need for repeated sampling -- to a small number of passes -- while still allowing effective updates to $\phi$ over several iterations. This can be particularly useful when $\mathcal{A}(\theta; \phi)$ involves expensive operations or when the gradient propagation is computationally intensive.

We also observed that it is beneficial to enforce the prior during the initial stages of optimization (by gradually scaling the DPS hyperparameter \(\zeta\) from \(0.1\) to \(1\)) and, later, to begin posterior sampling from an intermediate point—specifically, from a noisy version of \(\theta\) that retains some information, rather than from complete noise.

This approach allows $\theta$ to capture the global features early in the process, enabling the optimization to focus on refining other aspects of the rendering, such as finer details and complex scene parameters, in the subsequent steps. This strategy accelerates convergence and enhances the reconstruction's overall quality, helping avoid ambiguities and preventing premature convergence to local minima. 

Finally, an optional refinement step can be applied, which enforces data consistency \cite{song2023solving} before the latent $\theta$ is reused to improve $\phi$ and diffuse for the next step. This is achieved by directly optimizing the latent without any prior supervision. The rationale is that certain features will be preserved, allowing the latent to converge more quickly without constraints. Additionally, if ambiguity arises, it is advantageous for it to be reflected initially in the parameter that is subsequently ``cleaned'' by the prior. In practice we applied it a few steps around the middle of the process, to avoid early local minima in the beginning and artifacts due to overfitting at the end. The proposed optimization is outlined in Algorithm \ref{code:pdps}.

\begin{algorithm}
    \caption{Reconstruction with PDPS}
    \label{code:pdps}
    \begin{algorithmic}
\Require \State $y,\mathcal{R}, \mathcal{D}, \phi_0, \theta_0, p(\theta | y; \phi)$
\State $P$ \Comment{Number of passes}
\Statex
\State $\mathcal{L}(\phi, \theta) := \|y - \mathcal{R}(\mathcal{D}(\theta), \phi)\|_2^2+\mathcal{L}_{\textsc{reg}}(\phi)$
\For {$s = 1 \dots P$}
    \State $\phi_s \gets$ \Call{Optimize-$\phi$} {$\mathcal{L}, \phi_{s-1}, \theta_{s-1}$} \Comment{SGD}
    \State $\hat\theta_s \sim p(\hat\theta_s \, | \, y; \phi_s)$ \Comment {DPS}
    \If {$s \in S_\text{refine}$}
    \State $\theta_s \gets$ \Call{Optimize-$\theta$} {$\mathcal{L}, \phi_{s}, \hat\theta_{s}$} \Comment{Refinement}
    \Else
    \State $\theta_s \gets \hat\theta_s$
    \EndIf
\EndFor
\State \Return $\phi_P, \theta_P$
    \end{algorithmic}
\end{algorithm}

\section{Results}
\label{sec:results}

In this section, we demonstrate the effectiveness of our method for different use cases. All modules are implemented in Pytorch~\cite{paszke2019pytorch} and Vulkan SDK.  
Further details are provided in the supplementary material.
The code and the Cloudy dataset are publicly available at \url{https://www.github.com/rendervous/cloudy_project}.

\subsection{Diffusion Posterior Sampling}
In the first experiment, we shed light on the potential of DPS for single-view volume reconstruction. Through this experiment, we do not optimize for any physical parameters affecting the cloud appearance, but solely assess the strength of the volume diffusion prior when used to constrain the differentiable volume renderer. 

\begin{figure}[t]
    \centering
    \includegraphics[width=0.95\linewidth]{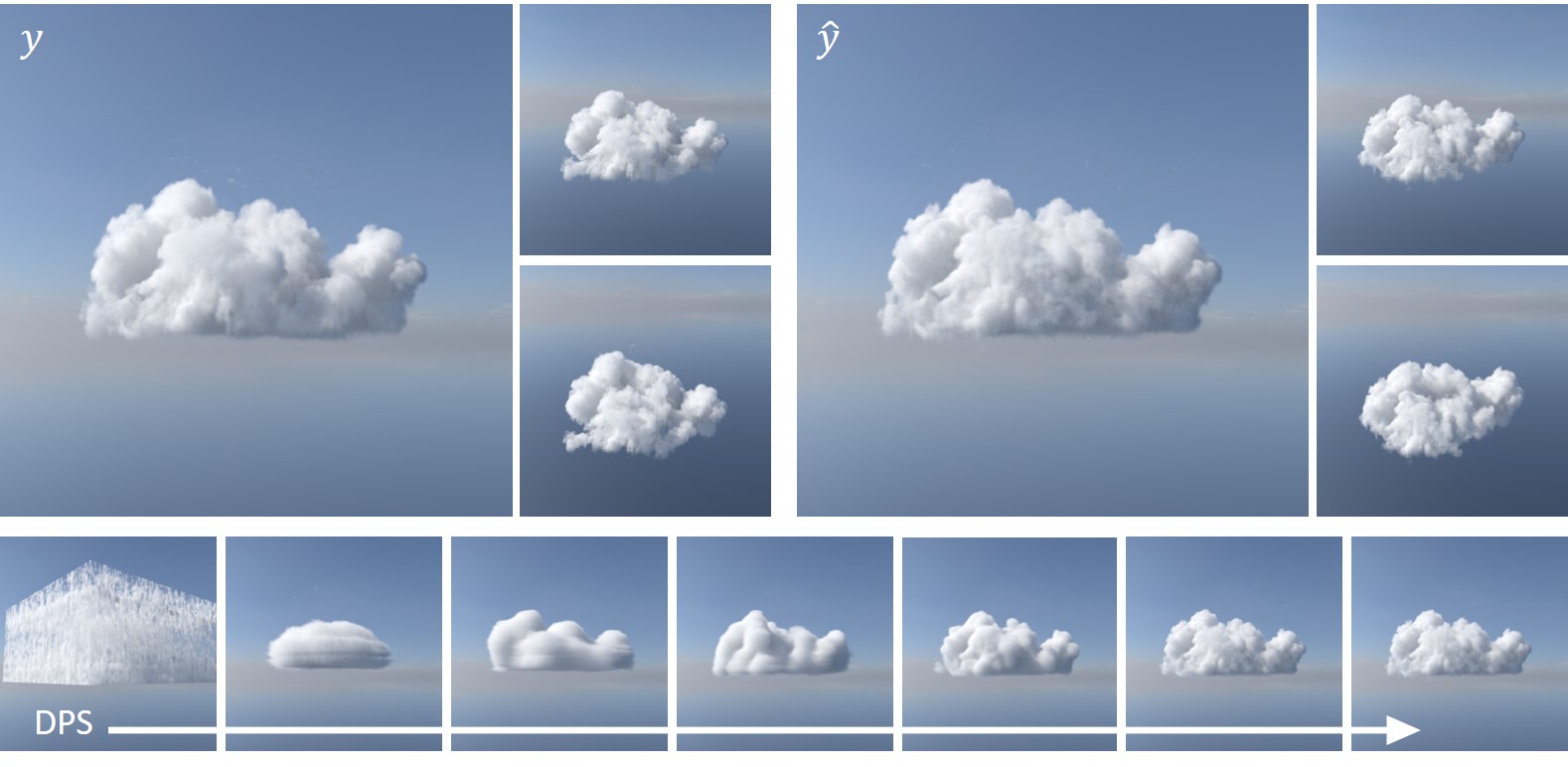}
    \caption{Diffusion Posterior Sampling. Given an observation and a differentiable process (differentiable volume rendering in our application), the denoising process is guided step-by-step toward matching the observation. From a different view, the reconstructed cloud may deviate from the ground truth, but the diffusion prior ensures that a realistic cloud is generated.}
    \label{fig:dps_example}
\end{figure}

Fig.~\ref{fig:dps_example} demonstrates this with a cloud from the Cloudy dataset, which is rendered with an environmental sky model and preset material properties. The Henyey-Grenstein scattering function approximation \cite{henyey1941diffuse} is used along with realistic values for the material absorption and scattering properties. More results are given in the supplementary material.  

The result shows how the denoiser is guided by the cloud's appearance, which is considered by the differentiable renderer, rather than performing unconditional denoising based solely on the diffusion model. Specifically, in each iteration, the current image-based loss is used to guide the sampling in the diffusion latent space. While an exact match with the given observation cannot be achieved -- since the denoiser cannot perfectly reproduce the corresponding 3D cloud -- the reconstruction fairly accurately matches both the observation (when rendered from the same view) and the 3D density field. Novel views of the reconstructed cloud and the ground truth further support the quality of our proposed single-view reconstruction.

\subsection{Monoplanar Representation}

To assess the quality that is achieved with the proposed monoplanar latent representation, we perform a series of experiments with the monoplanar, triplanar and dense grid representations. All representations use the same number of parameters for the latent, i.e.: Monoplanar $128\times 128 \times 32$, Triplanar $3\times 128 \times 128 \times 11$, and Grid $32 \times 32 \times 32 \times 16$. An upsampler is used in the cases of monoplanar and triplanar representation.

Table ~\ref{tab:rec_metrics} shows the average values for each metric across nine reconstructions using clouds from the Cloudy dataset. 
\begin{table}[h]
    \centering
    \small
    \begin{tabular}{ccccc}
         \textbf{Representation} & \textbf{PSNR$\uparrow$} & \textbf{RMSE$\downarrow$} & \textbf{MAE$\downarrow$} & \textbf{SSIM$\uparrow$} \\
         Triplanar & $38.13$ & $0.01245$ & $0.00417$ & $0.8547$ \\
         Grid & $37.26$ & $0.01377$ & $0.00436$ & $0.8412$ \\
         Monoplanar & $\textbf{38.46}$ & $\textbf{0.01199}$ & $\textbf{0.00393}$ & $\textbf{0.8609}$
    \end{tabular}
    \caption{Quality metrics for different latent representations.}
    \label{tab:rec_metrics}
\end{table}

While PSNR, RMSE, and MAE consider the full volume at $256\times128\times256$ resolution, SSIM~\cite{wang2004image} considers the center slice. Our proposed monoplanar representation quantitatively outperforms the other state-of-the-art representations in terms of reconstruction fidelity.

\begin{figure}[t]
    \centering
    \includegraphics[width=0.5\linewidth]{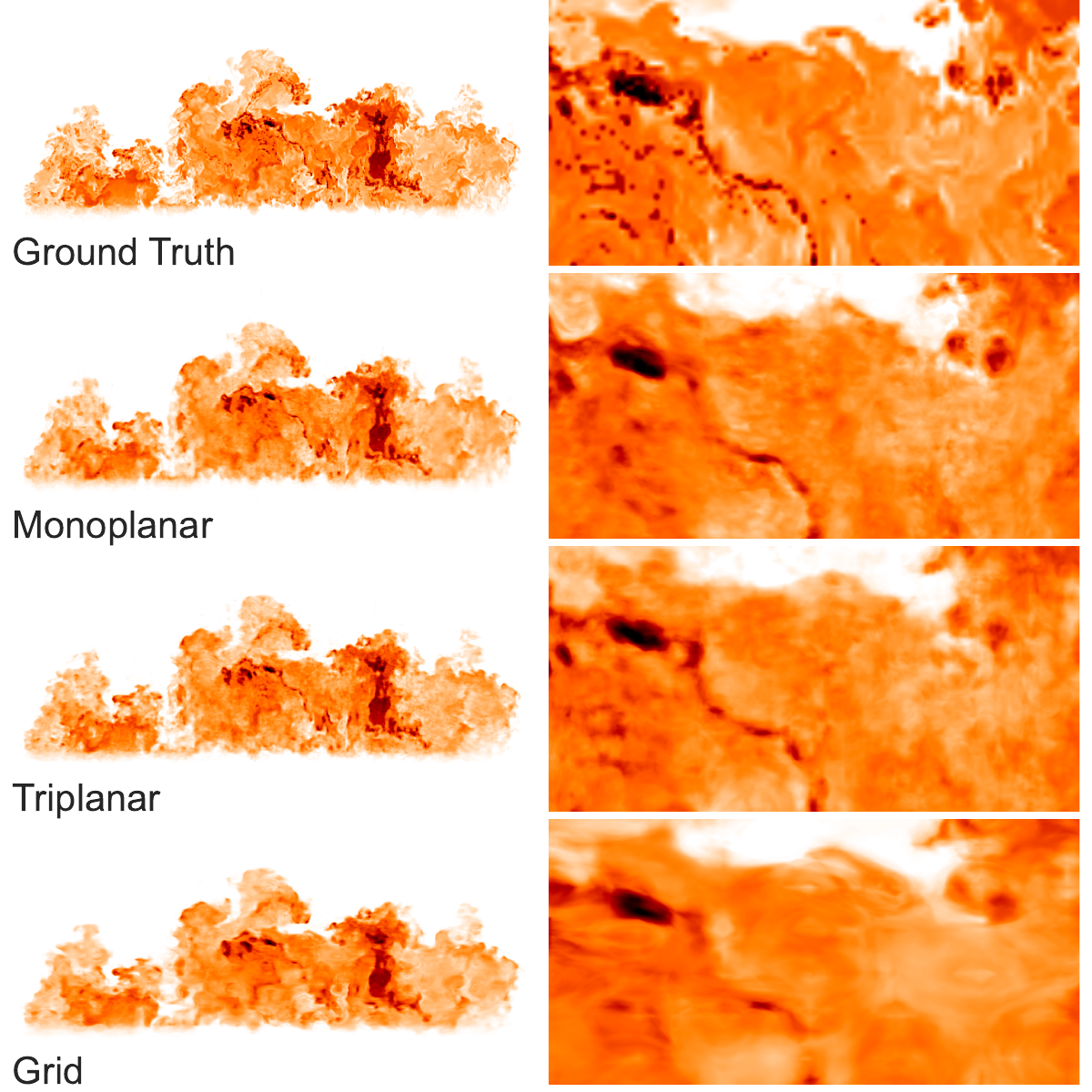}
    \includegraphics[width=0.48\linewidth]{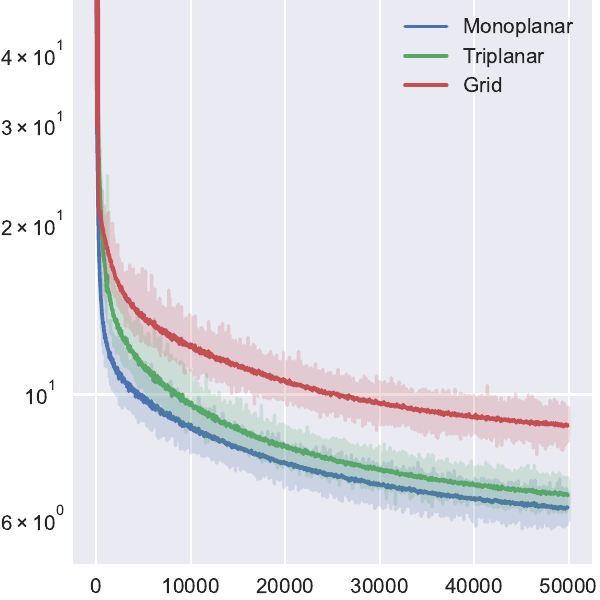}
    \caption{Qualitative comparison. Left: Cross-sections of a cloud and its reconstructions using different latent representations are shown. Right: Convergence graphs of the reconstruction loss over 50,000 steps, measured at 128K uniform sampled positions.}
    \label{fig:rec-qualitative}
\end{figure}

The qualitative comparison in Fig.~\ref{fig:rec-qualitative} highlights the strength of the monoplanar representation for volume reconstruction. Among all representations, features in the original cloud are best preserved, and the reconstruction loss for the monoplanar representation decays the fastest over the optimization iterations, decreasing monotonically toward the minimum.

\subsection{Super-Resolution}

Super-resolution is a common use cases for diffusion models. The diffusion process naturally integrates prior knowledge, making it effective in reconstructing fine details and completing structures in a plausible manner. 

\begin{figure}[t]
    \centering
    \includegraphics[width=.8\linewidth]{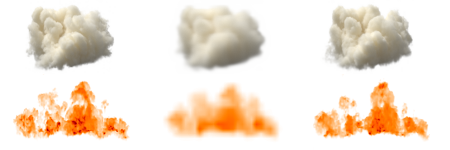}
    \caption{Cloud Super-Resolution. From a cloud on a $32\times16\times32$ grid (center), the diffuser reconstructs a density distribution on a $256\times128\times256$ grid (right). This process adds fine details and internal structures, demonstrating the model's ability to upscale and introduce complexity while preserving the overall coherence and shape of the original cloud (left).}
    \label{fig:cloud_super_resolution}
\end{figure}

For super-resolution, the measurement function is $\mathcal{A}(\theta) := \mathcal{C} ( \mathcal{D}(\theta) )$, where $\mathcal{C}$ is a coarse jittered sampling of the decoded grid $\mathcal{D}$. Figures \ref{fig:cloud_super_resolution} demonstrate the ability of our diffuser to perform super-resolution, by using DPS due to the non-linearity of the latent decoder. The non-linearity requires careful computation of the gradients with respect to $x_t$, to enable approaching a solution at $x_t$ that satisfies $y = \mathcal{A}(\hat{x}_0(x_t))$.

\subsection{Cloud Recovery from Transmittance Measures}

\begin{figure}[t]
    \centering
    \includegraphics[width=0.8\linewidth]{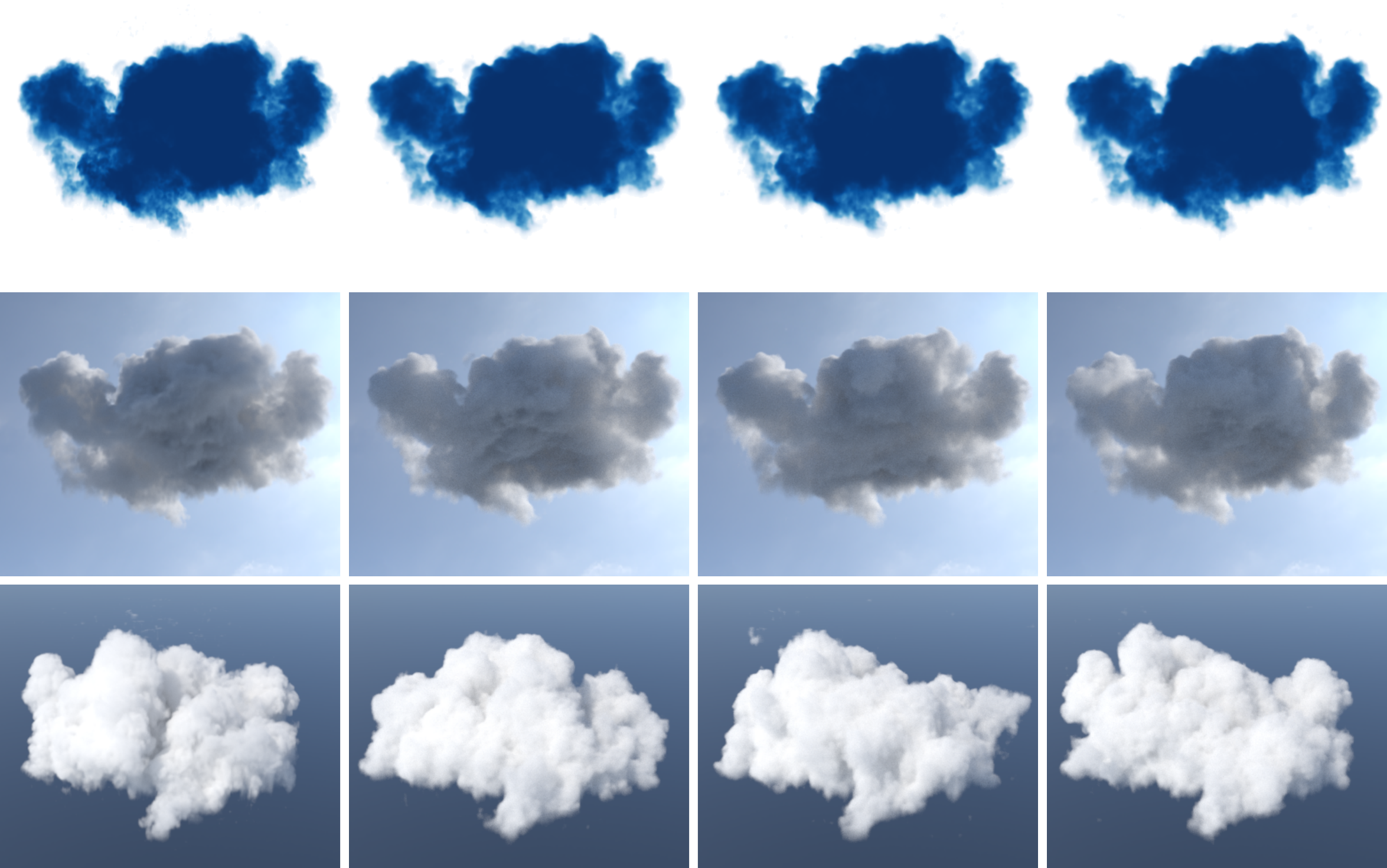}
    \caption{Transmittance-based single-view reconstruction. Left: Ground truth. The next columns show clouds conditioned on the transmittance image (top). Second row: Clouds rendered from the same view as the transmittance image. Third row: Novel views.}
    \label{fig:transmittance}
\end{figure}

DPS even has the capability to reconstruct a volume from a 2D transmittance image, with only posterior sampling (Figure \ref{fig:transmittance}). In this case, the transmittance is directly used as forward model, i.e., $\mathcal{A}(\theta) := \mathcal{T}(\theta).$ This enables, for instance, the use of microwave measurements of cloud particle density with weather and Doppler radar. 

\subsection{Comparative Evaluation}

\begin{figure*}[!h]
    \centering
    \includegraphics[width=\linewidth]{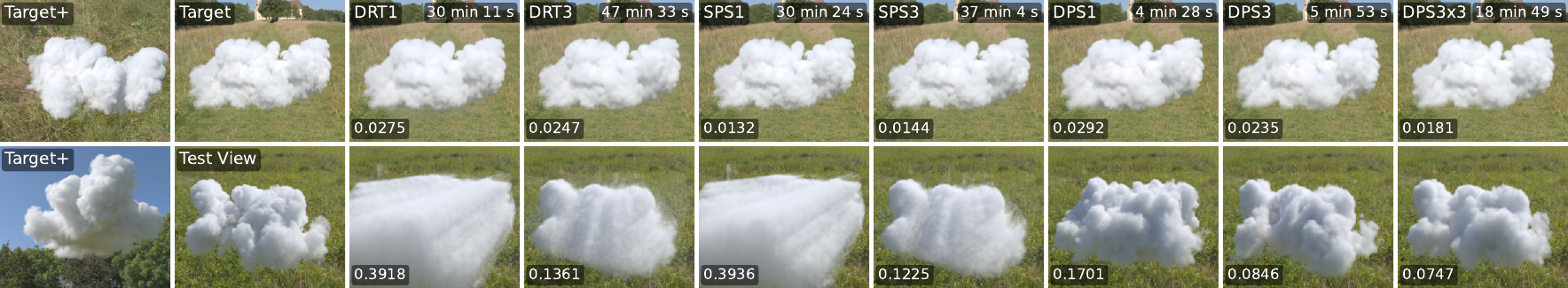}
    \caption{Reconstruction comparison. 
    The four leftmost \(2 \times 2\) images depict views used for reconstruction and testing. The number on the label indicates if 1 or 3 images were used for the reconstruction. The reconstruction time is reported in minutes (top), along with the LPIPS~\cite{zhang2018unreasonable} metric value (bottom), which quantifies the perceptual similarity between the ground truth and the synthesized views.
}
    \label{fig:few_view_rec}
\end{figure*}

\begin{table*}[h]
    \centering
    \small
    \begin{tabular}{c|c|c|c|c|c|c|c}
         \textbf{Metric} & \textbf{DRT1} & \textbf{DRT3} & \textbf{SPS1} & \textbf{SPS3} & \textbf{DPS1} & \textbf{DPS3} & \textbf{DPS3x3} \\
         T-LPIPS$\downarrow$ & 0.0323 & 0.0242 & 0.0123 & 0.0118 & 0.0205 & 0.0241 & 0.0124 \\
         N-LPIPS$\downarrow$ & 0.2937 & 0.1188 & 0.2869 & 0.1081 & 0.1126 & 0.0581 & 0.0572 \\
         Time & 00:30:40 & 00:40:58 & 00:31:44 & 00:33:54 & 00:03:44 & 00:04:23 & 00:14:47 \\
    \end{tabular}
    \caption{Quality comparison of DRT, SPS and DPS (ours) 
    using one and three views for reconstruction. The table shows average values over $32$ test cases, each constructed using clouds, materials, cameras, and environment settings sampled from 16 unseen clouds, 3 distinct cloud materials, 7 different environments, and 5 sets of camera poses.
    }
    \label{tab:stats}
\end{table*}

To compare our novel DPS approach with previous methods for reconstructing 3D clouds from images, we evaluate DPS alongside Differentiable Ratio-Tracking (DRT) 
\cite{nimier2022unbiased} and Singular Path Sampling (SPS) \cite{leonard2024image}. Since both DRT and SPS require multiple views to achieve accurate results, we tested with one and three images for the reconstructions.

We evaluate DPS under three different settings: (1) using only a single view (DPS1), (2) using all three views (DPS3), and (3) performing three restarts of the diffusion from a noisy version of a previously reconstructed latent (DPS3x3). The last setting aligns with diffuse-denoise strategies, progressively adjusting the initial noise toward the observed data to improve guidance stability. Results are shown in Fig.~\ref{fig:few_view_rec} and summarized in Table \ref{tab:stats}. 

The reconstructions using DRT and SPS show that while both techniques can overfit to a single view, they struggle to constrain unseen parts of the cloud, resulting in a smooth density distribution that only loosely follows the real distribution. By enforcing a prior on the cloud shape, as in DPS, we obtain a reconstruction in good agreement with the ground truth. Notably, even the single-view reconstruction aligns fairly well with the observed data, although challenges remain in capturing fine details.

\subsection{Recovering Light Conditions}

\begin{figure}
    \centering
    \includegraphics[width=\linewidth]{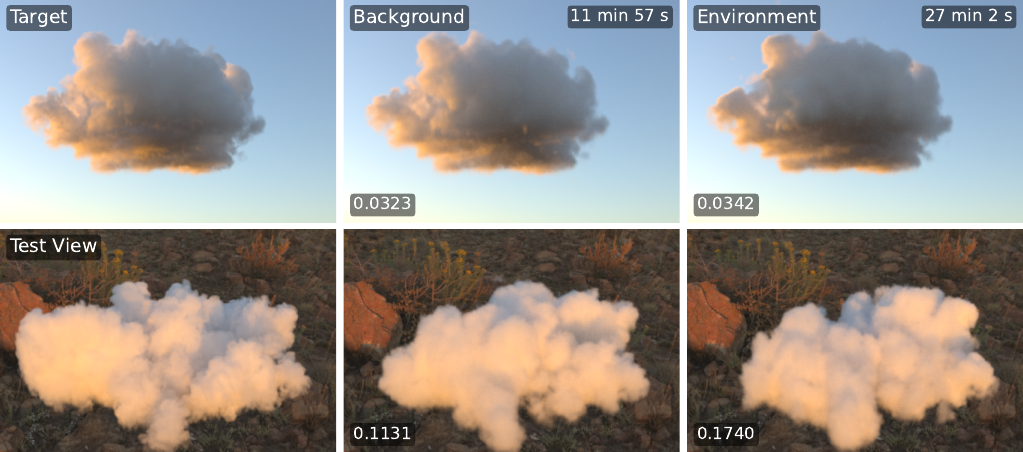}
    \includegraphics[width=\linewidth]{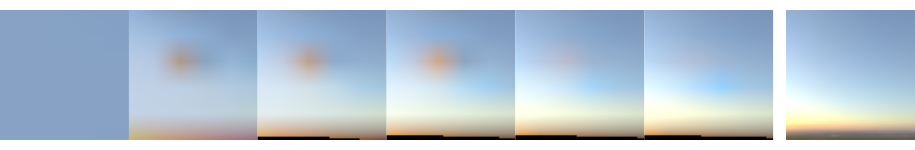}
    \includegraphics[width=\linewidth]{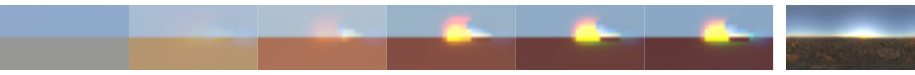}
    \caption{Recovering \(\phi\). Top: Reconstructions using parameterized DPS under two scenarios -- when the background radiance is unknown (Background), and when the entire lighting condition is unknown (Environment). Bottom: Evolution of the recovered background (top) and environment (bottom). Final column shows the lighting condition used to render the test views. }
    \label{fig:recovering_phi}
\end{figure}

Parameterized DPS is used in two scenarios: one where all physical parameters are known and the background needs to be recovered, and one where the entire lighting condition needs to be recovered (see Figure~\ref{fig:recovering_phi}). Despite the increasing complexity of each scenario, the reconstructions maintain consistent quality for both the target and novel views. Notably, the iterative optimization of lighting parameters for reproducing the test views converges to a setting that closely matches the one used to render these views.

\section*{Conclusions}
In this paper, we present a novel diffusion posterior sampling approach for single-view reconstruction of volumetric fields. Experimental results demonstrate that our approach provides robust generalization and achieves quality and performance that significantly exceed existing methods. With the availability of a few additional views, even more accurate reconstruction can be achieved.

A notable limitation is the ambiguity between what is represented by $\theta$ and $\phi$. For instance, background radiance may be misinterpreted as cloud structure, or parts of a cloud may be interpreted as 'painting' on the background radiance. If no proper regularization for $\phi$ is applied, the interleaved optimization of $\theta$ and $\phi$ may fall into local minima. This could lead to incorrect reconstructions, as certain parts of the cloud may be explained without actually being recovered. 

Further limitations arise from the use of a pre-trained diffusion model which, even for clouds alone, requires days to compute the latent encoding. Additionally, since a physically-based differentiable path tracer is employed to provide gradients, the reconstruction task is computationally intensive. This makes it challenging for our method to be applied to different phenomena such as smoke, fire, or explosions, and limits its use in time-critical reconstruction tasks, such as capturing time-varying phenomena. To address these issues, our approach may benefit from diffusion models trained specifically for direct 3D volume reconstruction from 2D images.
{
    \small
    \bibliographystyle{ieeenat_fullname}
    \bibliography{main}
}

\clearpage
\setcounter{page}{1}
\maketitlesupplementary

\section{Enhancing Latent Space}

\begin{figure}
    \centering
    \includegraphics[width=\linewidth]{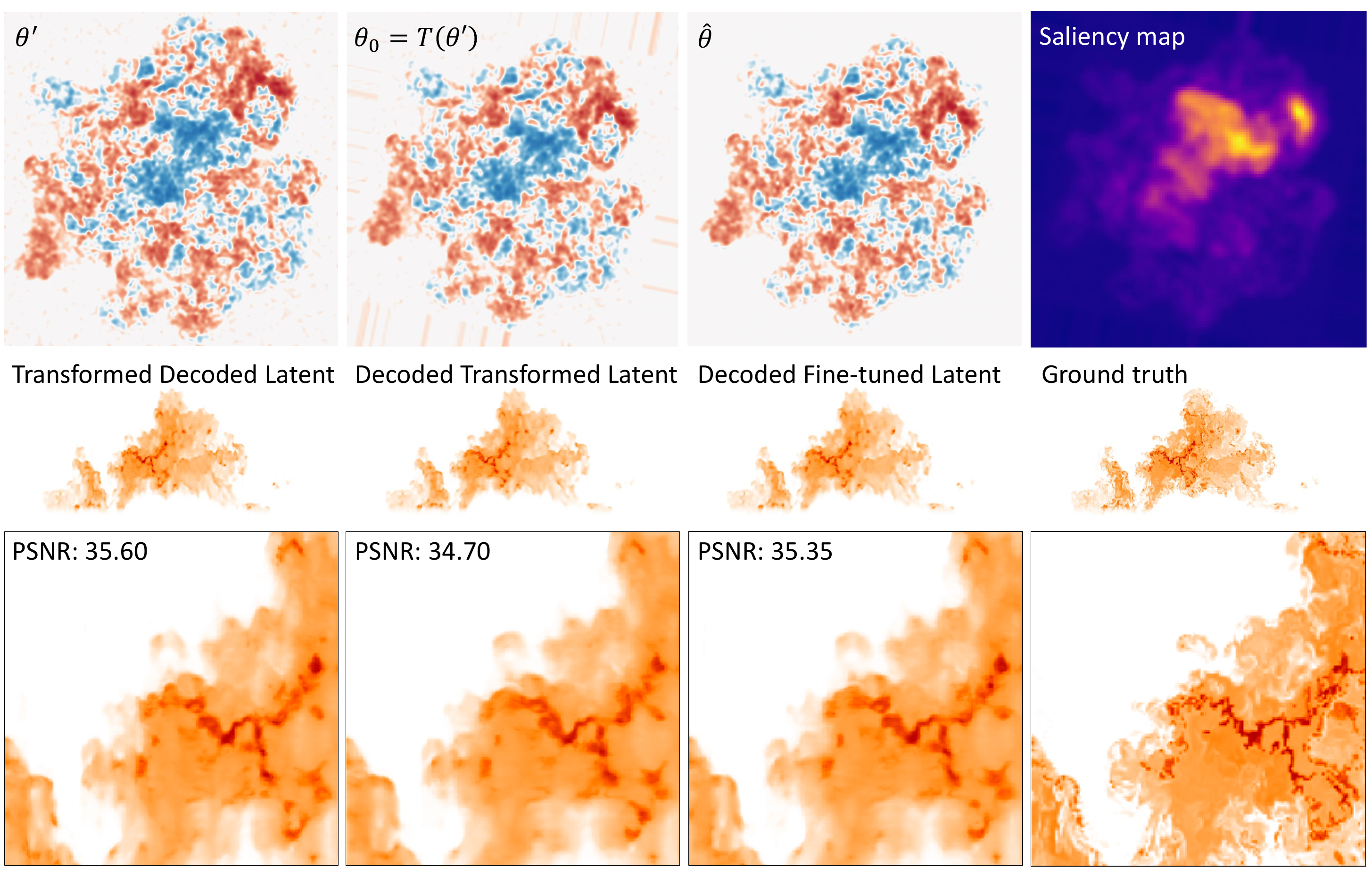}
    \caption{Latent enhancement. Starting with a latent code $\theta'$ obtained from the original volume, a transformed version serves as the initial solution $\theta_0$. A few optimization steps are performed to refine the latent representation $\hat{\theta}$, reducing artifacts and enhancing the peak signal-to-noise ratio (PSNR). During optimization, a saliency map derived from $\theta_0$ guides the process by adaptively sampling positions in regions with more prominent features.
 }
    \label{fig:latent_enh}
\end{figure}

Augmenting the original $1,000$ instances in the Cloudy dataset with additional volumes obtained via transformations 
requires increasing the encoding time significantly. For example, if encoding $1,000$ clouds requires \(2\) days on an NVIDIA GeForce RTX 3090, performing a \(14\)-fold multiplication would result in a total computational time of approximately one month.

We leverage the transformation consistency of our monoplanar representation with respect to the $xy$-plane. The key to reducing the encoding time from 2 minutes to approximately 12 seconds lies in initializing the latent code by applying the desired transformation directly to the original latent representation. Instead of evaluating the representation loss uniformly across all locations, we concentrate sampling in regions where features are most prominent, guided by a distribution derived from a saliency map. This approach uses the features of the initial solution, as the final solutions are expected to remain close to the initialization (see Figure~\ref{fig:latent_enh}).

Another benefit of this refinement is the reduction of patterns that typically emerge from clamping at the domain boundaries when sampling rotated or scaled positions. This helps prevent the generative model from misinterpreting those artifacts as valid structures.

\section{Differentiable Volume Rendering Module}

The rendering equation
assumes that light travels unchanged between visible surface positions, i.e., the incoming radiance at a point $x_a$ from $x_b$ remains unchanged; $L_i(x_a,\omega)=L_o(x_b, -\omega)$. However, incorporating participating media like clouds requires considering the interactions of light with particles within the volume, due to scattering and/or absorption effects (see Table~\ref{tab:symbols} for the notation used).

\subsection{Volume Rendering Equation}

\begin{table}
    \centering
    \begin{tabular}{p{0.22\columnwidth} p{0.65\columnwidth}}
        \hline
         \textbf{Notation} & \textbf{Description}  \\\hline
$\sigma_t(x)$ & \small{Extinction field, informally, the density distribution of the particles in the space.}\\
$\varphi(x)$ & \small{Scattering albedo: the probability of light to be scattered after a particle interaction.} \\
$\rho(\omega_i, \omega_o)$ & \small{Phase function: directional distribution of the scattered light.}\\
$B(\omega)$ & \small{Environment radiance coming from $\omega$.}\\
$T(x_a \leftrightarrow x_b)$ & \small{Transmittance between two positions.} \\
$L_s(x, \omega)$ & \small{Scattered light at $x$ towards $\omega$.}\\
$L_e(x, \omega)$ & \small{Emitted light at $x$ towards $\omega$.}\\
$L_i(x,\omega)$ & \small{Incoming radiance at $x$ from direction $\omega$.}\\
$L_o(x,\omega)$ & \small{Outgoing radiance at surface position $x$ towards direction $\omega$.}\\\hline
    \end{tabular}
    \caption{Terms involved in the volume rendering equation. Notice that all terms are wavelength-dependent.}
    \label{tab:symbols}
\end{table}

The \textit{Volume Rendering Equation (VRE)} computes the incoming radiance \( L_i(x_0, \omega) \) by integrating the contributions of scattered and emitted light along a ray, as well as direct contributions from surfaces. It accounts for transmittance (\(T\)), scattering properties (\(\sigma_t\), \(\varphi\), and \(\rho\)), and either volume emission or surface exiting radiance (\(L_e\) or \(L_o\)). 

Given the scattered radiance at \(x\) in the direction \(\omega\):
\[
L_s(x, \omega) = \int_{\omega_i} \rho(-\omega_i, \omega) L_i(x, \omega_i) \, d\omega_i,
\]
the incoming radiance at any point in space, including camera sensors, is computed as

\begin{equation}
\begin{split}
L_i(x_0, \omega) = & \int_0^d T(x_0 \leftrightarrow x_t) \sigma_t(x_t) \big[ \varphi(x) L_s(x, -\omega) \big. \\
& \big. + (1 - \varphi(x))L_e(x, -\omega) \big] \, dt \\
& + T(x_0 \leftrightarrow x_d) L_o(x_d, -\omega).
\end{split}
\label{eq:vre}   
\end{equation}

The recursive nature of equation \ref{eq:vre} is typically addressed using path sampling methods. In the path-based approach, a path \( z = x_0, \dots, x_N \) is sampled, where intermediate vertices correspond to scattering events and the final vertex represents either an absorption event or a surface interaction. The \textit{path throughput} \(\Gamma(z)\) captures the cumulative effects of transmittance, densities, scattering albedo, and phase functions along the path. In path-space, the expected radiance is expressed as
\[
L_i(x_0, \omega) = \int_{z} \Gamma(z) E(z) \, dz,
\]
where \(E(z)\) represents either volume emission (\(L_e\)) or outgoing surface radiance (\(L_o\)), depending on the final vertex. For simplicity, our analysis considers a single medium surrounded by a ``radiative environment shell'' that emits radiance inward (\(L_o(x, -\omega) = B(\omega)\)).

\textit{Volumetric path tracing} is a standard method for sampling paths proportional to \(\Gamma(z)\). However, in its basic form, this approach often experiences high variance due to a mismatch between the path throughput distribution \(\Gamma(z)\) and the radiance distribution of the environment. To address this, \textit{next-event estimation} reduces variance by considering direct contributions from the environment at each vertex along the primary path.

\subsection{Differentiable Rendering}

Let $\mathcal{R}$ be the process of computing the appearance of the volume $\mathcal{D}(\theta)$ subject to physical parameters $\phi$, by measuring the arriving radiance $L_i$ to an array of $W\times H$ sensors, i.e.,

$$
\mathcal{R}(\mathcal{D}(\theta); \phi) := \{ I_k \}_{k=1}^{W\times H}
$$
with $I_k = \int_{x_0,\omega} W_e^{(k)}(x_0,\omega) L_i(x_0,\omega) d{x_0}d{\omega}$. 
Here, \(x_0, \omega\) represents the incoming ray to the sensor, and \(W_e^{(k)}\) is a function that models the sensor's response, typically used to simulate complex lens optics or filter effects. The integral is approximated by averaging multiple samples per pixel, typically \(64\) in most cases. 

Since camera parameters (which could affect $W_e$ or the integral's limits) are not considered, derivatives of $\mathcal{R}$ with respect to its parameters propagate directly through the integral, i.e.:
$$
\partial_{\theta\phi} \mathcal{R}(\cdot)=\left\{ \int W_e^{(k)}(x_0,\omega)\partial_{\theta\phi}L_i(x_0,\omega) dx_0 d\omega \right\}_{k=1}^{W\times H}.
$$

The propagation of the gradients $\nabla_{\mathcal{R}}\mathcal{L}$ through all volumetric fields requires complex light-path sampling depositing the radiative quantities at every path interaction. 

\subsection{Differentiable VRE}

\begin{figure*}
    \includegraphics[width=0.33\textwidth]{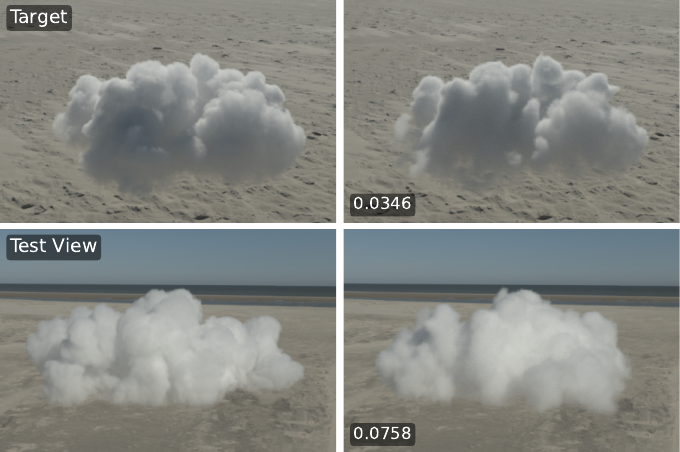}    
    \includegraphics[width=0.33\textwidth]{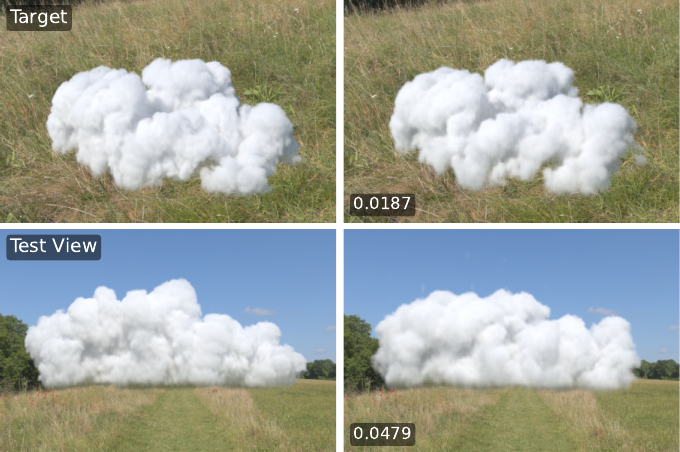}
    \includegraphics[width=0.33\textwidth]{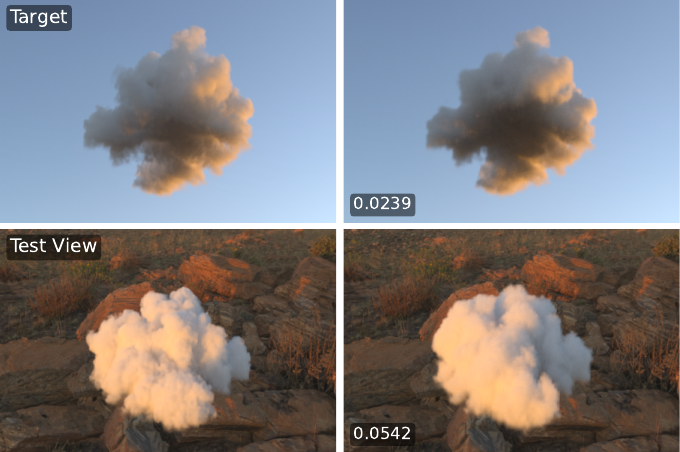}
    
    \includegraphics[width=0.33\textwidth]{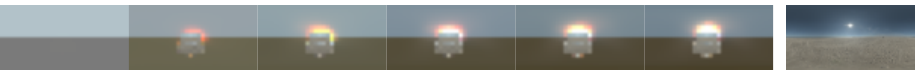}   
    \includegraphics[width=0.33\textwidth]{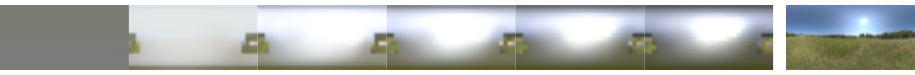}
    \includegraphics[width=0.33\textwidth]{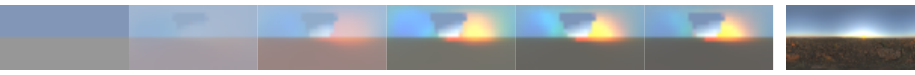}
    \caption{Additional results for reconstructions of both, cloud and lighting conditions, varying the material settings of the cloud and targeting different environment maps.}
    \label{fig:other_environments}
\end{figure*}

The propagation of gradients to the argument of an integral operator must adhere to the Leibniz Integral Rule. In this case, the integral limits are independent of the parameters, and there are no discontinuities in the fields. As a result, gradients with respect to \( L_i \) can be ``propagated'' directly to the integral argument. Specifically,

\[
\partial_{\theta\phi} L_i(x_0, \omega) = \int_z \partial_{\theta\phi} \left[ \Gamma(z) E(z) \right] \, dz.
\]

By applying the chain rule, the gradient of the loss function becomes

\[
\nabla \mathcal{L} = \int_z \nabla_{L_i} \mathcal{L} \cdot \partial_{\theta\phi} \left[ \Gamma(z) E(z) \right] \, dz.
\]

This is the idea proposed by Niemier et al. \cite{nimier2020radiative}, where path sampling is used to ``deposit'' gradients across all fields involved in the product \( \Gamma \). In \cite{vicini2021path}, the same \( z \) is replayed to compute both \( \Gamma \) and \( \partial \Gamma \). A tailored sampler \cite{nimier2022unbiased} is used to compute \( \partial_{\sigma(x_i)} \Gamma \), which becomes problematic when \( \sigma(x_i) \) is small. A weighted path sampler \cite{leonard2024image} includes singular paths with no more than one \( \sigma(x_i) = 0 \).

Summarizing, using techniques like DRT \cite{nimier2022unbiased} or SPS \cite{leonard2024image}, gradients with respect to the fields, such as \(\partial \mathcal{L} / \partial \sigma(x)\), can be computed. These fields may be represented using various spatial structures, including complex neural models. As long as the representations are differentiable, gradients can propagate to their underlying parameters. 

In practice, we use regular grids because they can be efficiently queried and are easily differentiable. If a more complex model is required, such as the volume decoder \(\mathcal{D}\), values at the grid vertices are evaluated to obtain the intermediate parameters \(\gamma\). Then, the gradients \(\nabla_\gamma \mathcal{L}\) are back-propagated through the model.

Finally, derivatives of \(\mathcal{R}\) with respect to \(\theta\) and \(\phi\) can be obtained using the differentiable volume renderer, and with this, the gradients of the loss function:
\[
\mathcal{L} = \| y - \mathcal{R}(\mathcal{D}(\theta), \phi) \|^2_2,
\]
that are required by the Diffusion Posterior Sampling and the \textsc{Optimization} method. In Fig.~\ref{fig:other_environments} we show some examples of the joint reconstruction of physical parameters $\phi$ (environment map) and density distributions of the cloud determined by $\theta$ with our proposed technique.

\begin{figure*}
    \centering
    \includegraphics[width=\linewidth]{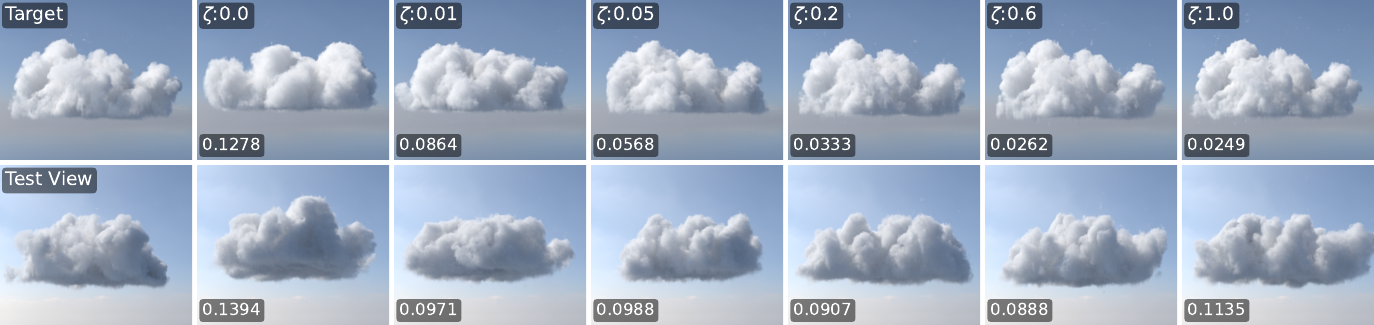}
    \caption{Effect of $\zeta$: Multiple DPS runs were performed with varying values of the $\zeta$ multiplier. The top row shows the reconstruction's approximation to the target view, while the bottom row presents the reconstruction from a different perspective. Higher $\zeta$ values lead to better alignment with the observation but deviate from the prior, resulting in less cloud-like formations. In contrast, smaller $\zeta$ values remain closer to the cloudy prior but exhibit weaker alignment with the observation.
 }
    \label{fig:dps_zeta}
\end{figure*}

\begin{algorithm}
    \caption{Parameterized DPS}
    \label{code:param_dps}
    \begin{algorithmic}
\Require \State $y, \mathcal{R}, \mathcal{D}, \phi$\
\State $\theta_{k}, k$ \Comment{Start noisy version}
\Ensure 
\State $\theta \sim p(\theta \, | \, y; \phi)$
\Statex
\For {$t = k \dots 1$} 
    \State $\boldsymbol{\epsilon} \gets \epsilon_\Phi(\theta_t, t)$  
    \State $\hat{\theta}_0 \gets \left(\theta_t - \sqrt{1 - \alpha_t} \boldsymbol{\epsilon} \right)/\sqrt{\alpha_t}$
    \State $\mathbf{z} \sim \mathcal{N}(\mathbf{0},\mathbf{I})$
    \State \Comment{DDIM step} 
    \State $\theta_{t-1}' \gets \sqrt{\alpha_{t-1}}\hat{\theta}_0 + \sqrt{1 - \alpha_{t-1} - \sigma_t^2}\boldsymbol{\epsilon}  + \sigma_t \mathbf{z}$
    \State \Comment{DPS step}
    \State $\theta_{t-1} \gets \theta_{t-1}' - \zeta_t \nabla_{\theta_t} \|y - \mathcal{R}(\mathcal{D}(\hat{\theta}_0), \phi)\|^2_2$ 
\EndFor
\State \Return $\hat{\theta}_0$
    \end{algorithmic}
\end{algorithm}

\section{Parameterized Diffusion Posterior Sampling}

Algorithm \ref{code:param_dps} outlines the adapted DPS method tailored for our parameterized posterior sampling approach. Here, \(\alpha_t\) denotes the noise scheduling parameter at time step \(t\). In practice, we sample only \(100\) time steps with a stride of \(10\), rather than sampling all steps. This adjustment also impacts the scaling factor \(\zeta_t\), which is proportionally amplified.

\subsection{Influence of $\zeta$ in DPS}

During diffusion posterior sampling, the gradients' scaling factor that guides the state toward the observation plays a crucial role in balancing the trade-off between prior enforcement and observation fidelity. The authors of \cite{chung2022diffusion} proposed the following formulation:

\[
\zeta_t = \frac{\zeta}{\| y - \mathcal{A}(\hat{x}_0(x_t)) \|},
\]

where the hyperparameter \(\zeta\) is chosen within the range \([0.1, 1.0]\). Figure~\ref{fig:dps_zeta} illustrates how this choice impacts reconstruction accuracy and adherence to the prior.

\section{Common diffusion-base tasks}

In this section, we present several applications of our proposed generative model and the parameterized diffusion posterior sampling technique, demonstrating their effectiveness across a variety of tasks. These applications highlight the versatility and power of our approach in addressing different challenges within the domain of volumetric scene reconstruction and rendering.

\begin{figure*}[ht]
    \centering
    \includegraphics[width=\linewidth]{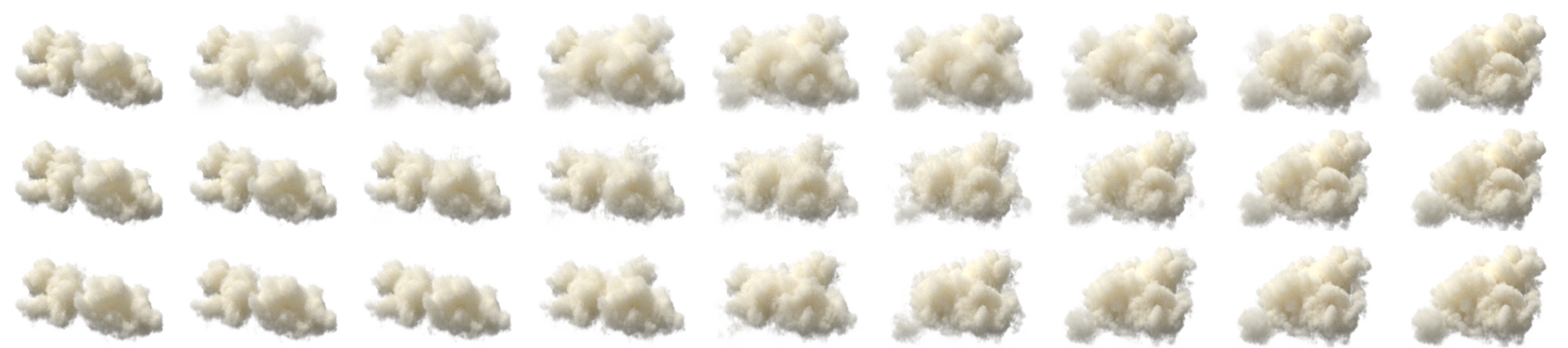}
    \caption{Cloud Interpolation. Top row: linear interpolation between grids, showing a straightforward blending of two cloud structures. Middle row: Linear interpolation between latent representations, offering smoother transitions compared to direct grid interpolation, but still revealing limitations such as ghosting effects. Bottom row: DPS (Diffusion Posterior Sampling) using the linear interpolation in latent space as the target, resulting in more coherent and natural transitions, with the prior enforced to avoid artifacts like ghosting.}
    \label{fig:dps_interpolation}
\end{figure*}
\subsection{Generative model}

One notable property of our proposed DDPM is its ability to generate new clouds. The generated clouds look similar to the original clouds in Cloudy, and their internal structure closely resembles that of a physical simulation.
This is demonstrated in Fig. \ref{fig:generating} in the main document.

\emph{Interpolation}: Interestingly, linear interpolation in the cloud's latent space—i.e., between different latent representations—produces plausible transitions between cloud shapes. However, when the cloud distributions differ significantly in terms of lobes or fine elongations, ghosting effects may occur as structures fade out linearly.

To address this issue, we propose an interpolation method based on posterior sampling: The mixture in the latent representation serves as the target, defined as $y := (1 - \alpha)\theta_a + \alpha \theta_b$, where $\theta_a$ and $\theta_b$ are the latent representations of two different clouds, and $\alpha$ controls the blending factor. This method ensures smoother transitions by taking the cloud structure into account during the interpolation process, and enforcing the prior to prevent the appearance of ghost artifacts. By integrating posterior sampling, the model adapts to the natural distribution of clouds, resulting in more physically consistent transitions.

Figure \ref{fig:dps_interpolation} showcases the differences between the linear interpolation strategy and our proposed method, highlighting the improved transitions and the reduction of ghosting effects in complex cloud distributions.

\subsection{Super-resolution and In-painting}

\begin{figure}[ht]
    \centering
    \includegraphics[width=\linewidth]{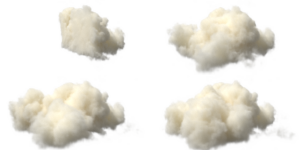}
    \caption{Cloud Inpainting. The diffuser is employed to generate a cloud that is consistent with a visible portion of the cloud. Three different instances are generated and displayed, demonstrating the model's ability to generalize and create diverse cloud formations, each unique yet adhering to the visible parts provided. }
    \label{fig:cloud_inpainting}
\end{figure}
 
Super-resolution and in-painting are common use cases in image restoration with diffusion models. These tasks are particularly well-suited for diffusers because the denoiser can easily preserve parts of the existing signal while filling in missing or low-resolution regions with consistent and coherent information. The diffusion process naturally integrates prior knowledge, making it effective at reconstructing fine details and completing structures in a visually plausible manner. 

For the case of super-resolution, our measurement function is $\mathcal{A}(\theta) := \mathcal{C} ( \mathcal{D}(\theta) )$, where $\mathcal{C}$ is a coarse jittered sampling of the decoded grid $\mathcal{D}$. In the case of in-painting, we assume a mask of interest $M$ and consider $\mathcal{A}(\theta) := M \otimes \mathcal{D}(\theta)$.

Figures \ref{fig:cloud_super_resolution} and \ref{fig:cloud_inpainting} demonstrate the performance of our diffuser on super-resolution and in-painting tasks respectively. While these tasks are typically linear in explicit cases, we continue to use Diffusion Posterior Sampling (DPS) due to the non-linearity of our latent decoder. This non-linearity complicates the optimization, and therefore approaching the solution at $x_t$ to satisfy $y = \mathcal{A}(x_0(x_t))$ requires careful computation of the gradients with respect to $x_t$.

\section{Extended comparisons}

\begin{figure*}
    \includegraphics[width=\linewidth]{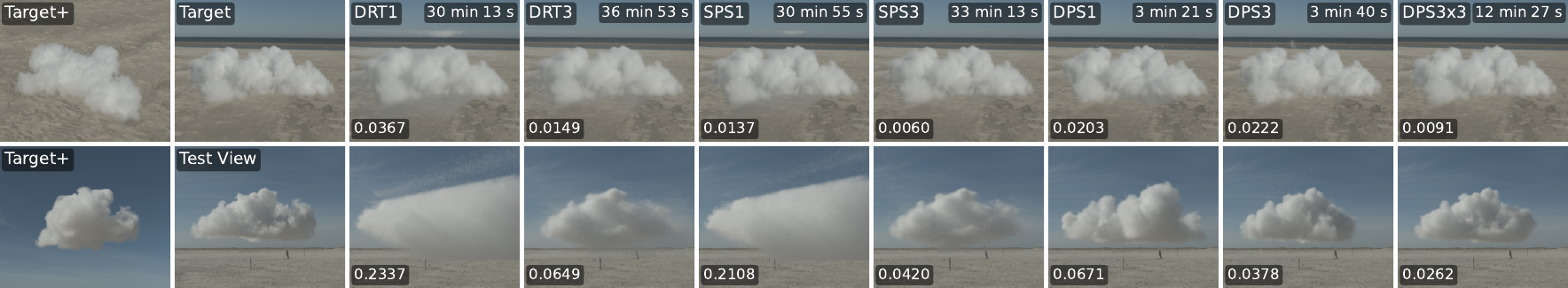}\\

    \includegraphics[width=\linewidth]{sec/images/comparisons_rendered_2_2_1_2.pt.pdf}\\

    \includegraphics[width=\linewidth]{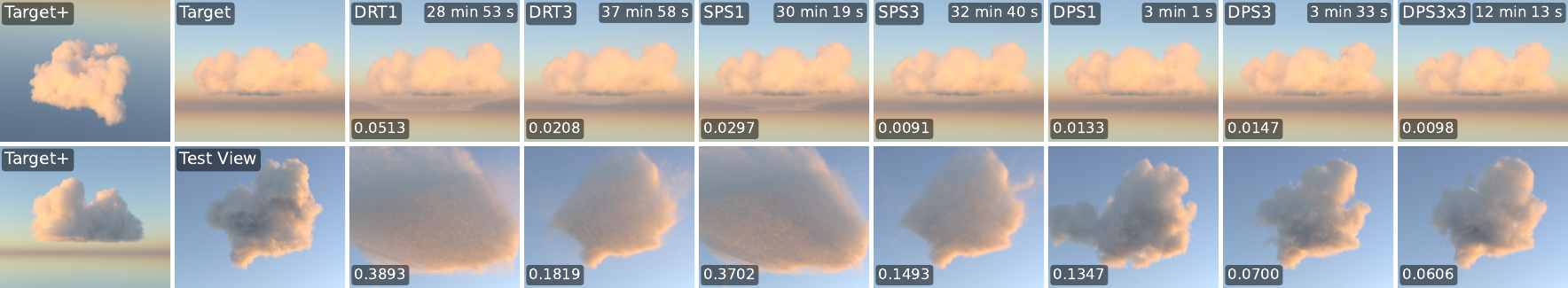}\\

    \includegraphics[width=\linewidth]{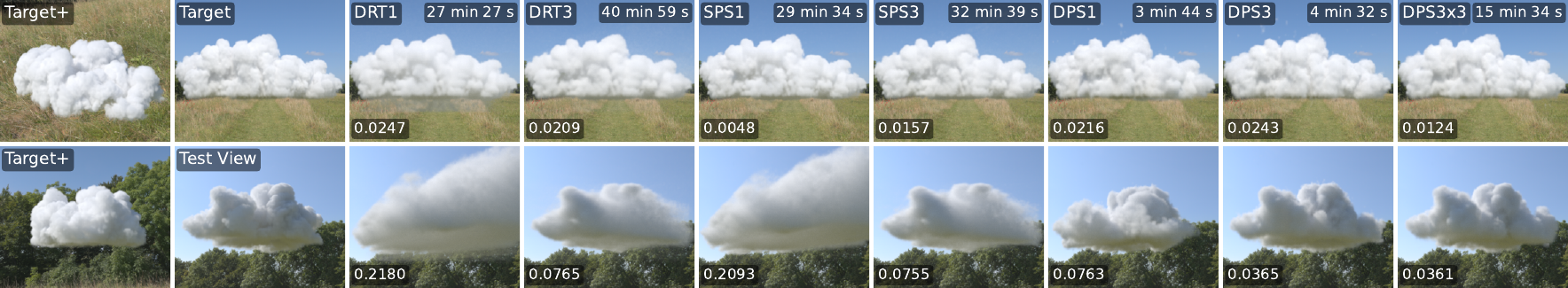}\\

    \includegraphics[width=\linewidth]{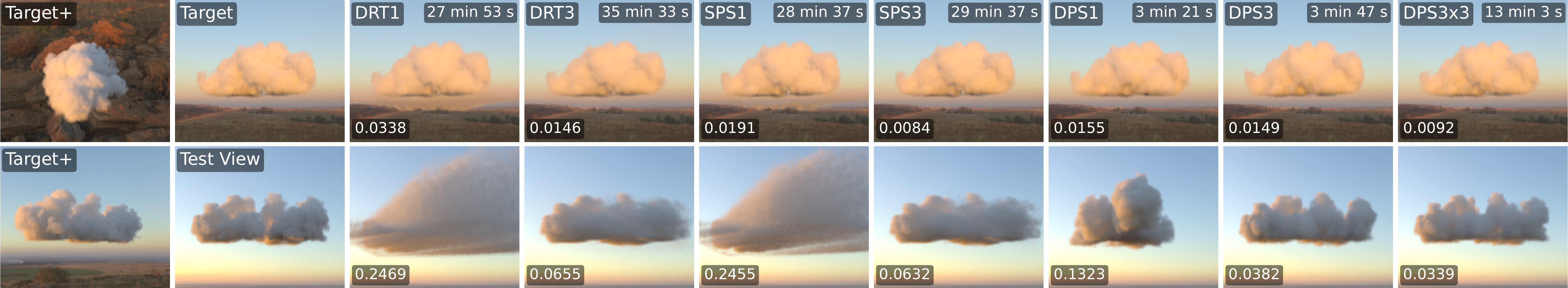}\\

    \caption{Further comparisons between different reconstruction techniques for single- and sparse-view settings.}
    \label{fig:extended_comparisons}

\end{figure*}

Fig.~\ref{fig:extended_comparisons} shows visual examples from the 32 test cases.

\end{document}